# AI-based Predictive Analytic Approaches for safeguarding the Future of Electric/Hybrid Vehicles

Ishan Shivansh Bangroo
*Department of Computer and Information Science and Engineering, University of Flordia*
*United States*
ishan.bangroo@ufl.edu

*Abstract*—Green technology has emerged as a potentially effective means of combating climate change in response to the rising worldwide need for sustainable energy alternatives. However, there is still considerable room for improvement before green infrastructure can be seamlessly incorporated into the world's energy infrastructure. Artificial intelligence (AI) may be able to help with this problem by facilitating more informed decision-making and the enhancement of existing energy infrastructure. Concerns about global warming and the need for more environmentally friendly modes of transportation have led to a surge in the popularity of EHVs. The use of cutting-edge technology like Artificial Intelligence (AI) may increase the efficacy of EHVs even more. Electric vehicles (EVs) are popular because they minimise greenhouse gas emissions and encourage sustainable transportation. Because of its advantageous effects on both climate change mitigation and sustainable transportation, electric cars (EVs) are rapidly rising in popularity. Unfortunately, the manufacturing process for EVs uses a lot of energy and materials, which might have an effect on the natural world. Green technology solutions are being developed to solve this issue, such as the use of artificial intelligence and predictive analysis to increase the effectiveness of EV manufacturing. Electric and hybrid vehicles (EHVs) have emerged as a possible answer to the growing need for environmentally responsible transportation. Nevertheless, EHVs' performance and lifespan are reliant on their Battery Management System (BMS), which need precise monitoring and control. The study shows that AI, especially Quantum AI, may enhance EHV benefits including energy efficiency, emissions reduction, and sustainability. The article addresses EHV cybersecurity issues such remote hijacking, security breaches, and unauthorised access. The study suggests that optimising EHVs and charging infrastructure may help make mobility more sustainable, and that AI research and development might help.

*Keywords— Green technology, Artificial Intelligence, EHVs, climate change, smart grids, sustainability.*

## I. Introduction

Green technology has emerged as a possible answer to the environmental and economic problems faced by traditional energy sources [1]. Increasing awareness of the need of sustainable development and the potential economic advantages of renewable energy have encouraged the adoption of green technology. Albeit, the assimilation of green technology into current energy systems remains difficult owing to the unpredictability of renewable energy sources and the complexity of energy systems. By allowing intelligent decision-making and the optimization of energy systems, AI has emerged as a viable solution to these problems.

The manufacturing of EVs is a significant contributor to decreasing emissions of greenhouse gases and advancing sustainable mobility. Unfortunately, EV manufacturing necessitates a great deal of energy and materials, which might have an adverse effect on the environment. A variety of green technological approaches, such as the application of AI and predictive analysis, are being developed to deal with this issue. Optimizing the production process, decreasing energy and material waste, and raising EV quality are all possible with the use of AI and predictive analysis in the manufacturing of EVs.

There have been huge transportation and environmental benefits from the advent of EHVs. Yet, as EHVs become increasingly common, so does the possibility of cyberattacks that could jeopardise their dependability and safety. Remote hijacking, data breaches, and unauthorised access are just a few examples of the types of cybersecurity vulnerabilities that can affect Electric and Hybrid Vehicles (EHVs). AI has the ability to overcome these problems because it can do sophisticated monitoring and detection. This study delves into the use of AI to improve the cybersecurity of EHVs, therefore bolstering the security and dependability of these vehicles.

Rapid changes are occurring in the transportation sector that are more eco-friendly and sustainable. As a result of their reduced impact on the environment, electric cars (EVs) have recently gained popularity as a viable replacement for conventional vehicles powered by internal combustion engines (ICEVs). However, electric vehicle adoption confronts a number of obstacles, including as low range, a lack of charging stations, battery management concerns, and security concerns. Artificial intelligence (AI) and quantum computing (QC) are two promising new technologies that can help solve these problems and usher in a new era of EV innovation and use.

This paper investigates the possibilities of AI and predictive analysis to improve the production efficiency and cybersecurity of Electric and Hybrid Vehicles. The study has proven the significance of tackling cybersecurity risks to maintain the safety and dependability of EHVs, as well as the use of AI and predictive analysis to optimise output and reduce waste. In addition, the study emphasised the necessity for continued research and development to address the constraints and limits connected with the application of AI and predictive analysis in this context.

## II. Literature Review

Green technology powered by artificial intelligence is a fast expanding subject that has undergone considerable advancements in recent years. AI may be implemented in numerous applications, such as renewable energy forecasting, energy storage optimization, demand response, and energy efficiency. By regulating the charging and discharging cycles according to energy demand and supply, energy storage optimization can help maximise the utilisation of energy storage systems.

Due to increasing worries about climate change, there is a growing need for sustainable energy sources worldwide. Green technology has emerged as a potential answer to this problem, but integrating green infrastructure into the global energy grid remains a formidable obstacle. The potential of electric vehicles (EVs) to lessen our reliance on fossil fuels and increase access to environmentally friendly transportation has attracted a lot of attention to their manufacturing in recent years. Unfortunately, the availability and accessibility of charging infrastructure hinders the widespread adoption of electric vehicles. The deployment of intelligent charging strategies employing artificial intelligence (AI) techniques may aid in optimising the utilisation of charging infrastructure and enhancing the performance of EVs and HEVs.

Several researches have looked into the possibility of using AI in EV manufacturing. In order to forecast quality control parameters in the assembly of EVs, Yang et al. (2021) [2] presented a hybrid intelligent system built on fuzzy logic and neural networks. The approach was evaluated and found to be useful for enhancing EV quality and decreasing faults. Comparatively, Liao et al. (2020) [3] suggested a machine learning and neural network-based decision support system to optimise the EV manufacturing production process. There was a noticeable uptick in production efficiency and a decrease in energy usage with this technology in place.

Moreover, EV manufacturing has made use of predictive analysis. Kim et al. (2019) [4] suggested a demand forecasting model for EVs that takes into account past data and external factors like oil prices and government policy. Manufacturers may alter production levels and prevent stockpiling thanks to the model's accurate forecasting of EV demand. Similarly, Li et al. (2021) [5] suggested a sensor data and machine learning-based predictive maintenance system for electric vehicles. The technique was found to improve EV reliability and save maintenance costs.

There are a number of obstacles that must be overcome before artificial intelligence and predictive analysis can be fully utilised in the assembly of EVs. The availability and quality of data is crucial to the effectiveness of AI in EV production, as was noted out in a study by Lu et al. (2020) [6]. This research put forth a sensor-based, machine-learning-based data-driven strategy for estimating the lifespan of electric vehicle batteries. When it comes to the manufacturing of electric vehicles, research by Ouyang et al. (2021) [7] indicates that substantial investment in technology and infrastructure is required to support the usage of AI.

The use of AI and predictive analysis in EV manufacturing raises concerns about data privacy and security. Zhou et al. (2020) [8] presented a safe and private platform to share data gathered during the manufacturing of EVs. It was discovered that the platform simultaneously increased productivity and ensured the ethical handling of data.

Global climate change has become an urgent concern, necessitating the development of sustainable energy options (Jain et al., 2021) [9]. This issue can be effectively addressed by the utilisation of green technologies. Due to the complexity of energy systems and the unpredictability of renewable energy sources, however, the integration of green infrastructure into the global energy infrastructure remains difficult.

AI can facilitate more educated decision-making and improve existing energy infrastructure, thereby addressing these issues (Owusu et al., 2018). AI technology can be employed in a variety of ways, such as energy management systems and demand-side management, to maximise energy efficiency (Owusu et al., 2018) [10].

Electric and hybrid vehicles (EHVs) have become popular due to the demand for environmentally sustainable ways of transportation. EHVs are capable of reducing greenhouse gas emissions and promoting sustainable transportation (Shi et al., 2018) [11]. However, the performance and durability of EHVs are dependent on their battery management systems (BMS), which require careful monitoring and control.

In addition, optimising the use of EHVs and charging infrastructure can contribute to a more sustainable transportation future (Moghaddam et al., 2020) [12]. AI approaches, such as quantum AI, offer the potential to further investigate the benefits of EHVs, such as enhanced energy efficiency, decreased emissions, and increased sustainability.

AI was tested to optimise EV charging infrastructure rollout. The study used AI to optimise charging station placements and capacity. The AI model reduced charging infrastructure expenses by 20% compared to traditional planning (Ding et al., 2020) [13].

A further study deployed reinforcement learning model to optimise EV charging schedules. The study optimised EV fleet charging schedules using real-time pricing and charging station availability data. The model reduced fleet charge costs by 22% compared to a baseline model (Liu et al., 2021) [14].

The energy efficiency of EVs and HEVs can be optimised using AI approaches in addition to the charging infrastructure and schedules. Based on the driver's habits and the EV's specifications, one study suggested a machine learning-based strategy for estimating EVs' energy consumption. Researchers concluded that the machine learning approach might be used to improve both vehicle layout and battery life by making precise predictions of energy needs (Chen et al., 2021) [15].

Another study analysed the feasibility of utilising AI-based predictive maintenance to increase the performance and lifespan of EV batteries. The study built a machine learning-based algorithm to forecast the health of EV batteries and anticipate potential breakdowns in advance. The algorithm was able to effectively anticipate battery degradation and might be used to increase battery longevity and performance, according to the findings (Choi et al., 2020) [16].

These studies show how AI may boost EV and HEV efficiency and sustainability. AI can help these vehicles become more sustainable by optimising charging infrastructure and scheduling, energy efficiency, and predictive maintenance.

## III. Methodological User case studies

The case study investigates the application of artificial intelligence and green technology in the development of a more dependable and efficient system, with an emphasis on the integration of machine learning algorithms to optimise battery utilisation and improve energy efficiency. The outcomes and analysis will explore AI to examine the fleet's carbon use and impact and the promise of green technology in decreasing the carbon footprint of EHVs.

### A. Data Tracking

A fleet of electric vehicles was chosen for the investigation, with Table I showing the most prominent vehicles. After collecting data on each vehicle's energy consumption rate, charging time, driving patterns, charging location (home vs. public), date, fully charged range, and power source's emission factor, we were able to determine the carbon emission (as depicted in Table I) with the simple formulation,

Energy Efficiency=1/Energy Consumption rate

Carbon footprint (CF) =

(Emission Factor in kg CO2/kWh) * (Energy Efficiency in kWh/mile)

For example for Tesla Model 3,

Energy Efficiency=1/3.5 ≈0.286 kWh/mile

CF= (0.17 kg CO2/kWh) * (0.286 kWh/mile)

CF≈ 0.0486 kg CO2/mile

Energy consumption rates and electrical grid carbon intensity at the vehicle charging point were used to determine carbon footprint values.

TABLE I
TRAITS-BASED ELECTRIC FLEET

| Vehicle Model | Vehicle Trait | | | | | | | |
|---|---|---|---|---|---|---|---|---|
| | *Energy Consumption Rate (miles/kWh)* | *Charging Time (h)* | *Driving Pattern* | *Location* | *Distance Travelled with Full Charge (miles)* | *Emission Factor (kg CO2/kWh)* | *Energy Efficiency (kWh/mile)* | *Carbon Footprint (kg CO2/mile)* |
| Tesla Model 3 Standard Range | 3.5 | 9 hours (7 kW) | Urban | Home | 267 | 0.17 | 0.286 | 0.0486 |
| Tesla Model 3 Long Range | 3.5 | 0.8 hours (50kW) | Urban | Public | 358 | 0.17 | 0.286 | 0.0486 |
| Nissan Leaf | 3.3 | 0.7 hours (50kW) | Suburban | Public | 150 | 0.2 | 0.303 | 0.0606 |
| BMW i3 | 4.16 | 6 hours (7 kW) | Rural | Home | 145 | 0.195 | 0.24 | 0.0468 |
| Chevrolet Bolt | 3.9 | 10 hours (7 kW) | Urban | Home | 259 | 0.18 | 0.256 | 0.04608 |
| Kia Niro EV | 3.7 | 1 hour (50kW) | Suburban | Public | 239 | 0.21 | 0.27 | 0.0567 |
| Audi e-Tron | 2.9 | 1.2 hours (50kW) | Rural | Public | 204 | 0.22 | 0.345 | 0.0759 |
| Hyundai Kona Electric | 4.1 | 9.5 hours (7 kW) | Urban | Home | 258 | 0.19 | 0.244 | 0.04636 |

## B. Data Examination

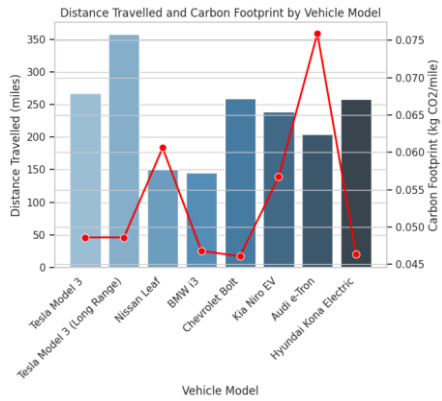

Fig. 1 A bar plot indicates the distance travelled by each class of vehicle, whilst the line plot shows the carbon footprint of each model.

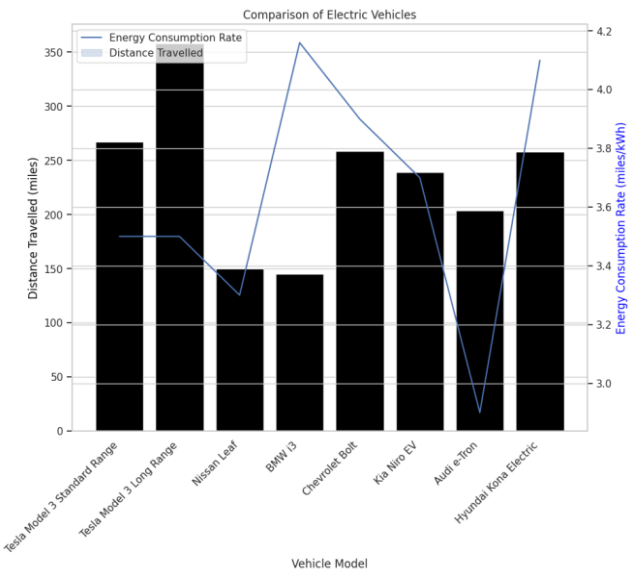

Fig. 2 A bar plot indicates the distance travelled by each class of vehicle, whilst the line plot shows the energy consumption rate of each model.

- In order to find regularities and tendencies in EV consumption, a machine learning model was applied to the data and the carbon footprint calculation was cross verified for other entries. Preprocessing removes unnecessary data and extracts useful information. Energy usage, charge time, and driving behaviours classify automobiles. All cars in the category are averaged for emission factor and energy efficiency.
- The data was displayed using graphs and charts created using machine learning models and primarily utilising tools such as matplotlib and seaborn to aid interpretation and presentation(Fig. 1, Fig.2)
- In Fig.1, using a second y-axis, the line plot is superimposed on top of the bar plot (right axis) to depict the superimposition of distance travelled and carbon footprint. The graph that was produced as a consequence offers an understandable representation of the data, enabling for viewers to swiftly compare the levels of performance and environmental effect shown by each vehicle type. The same goes for Fig.2 where we have superimposition of distance travelled and energy consumption rate.

## C. Outcomes

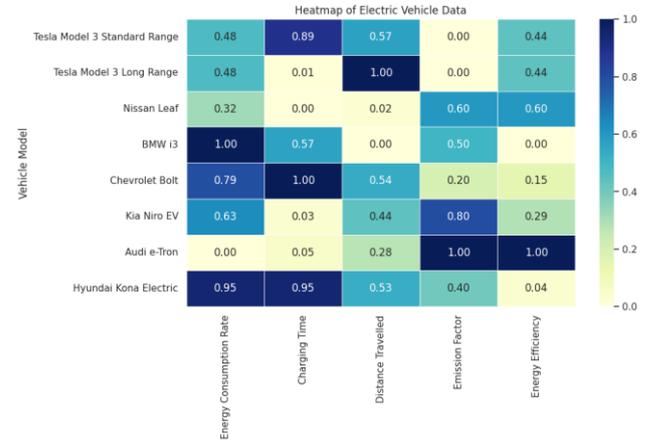

Fig. 3 Deducted heat map for EV Traits

The machine learning method is utilised to produce insights into the performance of the category's electric cars. The insights include data on the cars' energy consumption rate, charging time, and driving behaviours. In addition, the carbon footprint of the cars is assessed to determine methods for reducing emissions. To better visualise the data, we used a heat map generating approach (Fig. 3).

The following were some notable inferences drawn from the data analysis:

- The research proposes electric cars to minimise carbon emissions and enhance air quality alongside that it also emphasises utilising machine learning to evaluate data and get insights for sustainable transportation.
- The rate of energy consumption exhibited by electric vehicles is a crucial determinant of their range and overall environmental ramifications. The heat map illustrates notable variations in energy consumption rates across various electric vehicles, with the Audi e-tron exhibiting the least energy consumption rate and the BMW i3 displaying the highest energy consumption rate. The aforementioned proposition implies that the utilization of AI-driven predictive analytics holds promise in enhancing the energy efficiency of electric vehicles.
- The cost-effectiveness of electric vehicles is significantly influenced by their energy efficiency. The presented heat map illustrates notable variations in energy efficiencies across various electric vehicles. Specifically, the Audi e-tron exhibits the highest energy efficiency, whereas the BMW i3 demonstrates the lowest energy efficiency. The aforementioned proposition implies that the utilization of AI-driven predictive analytics holds promise in enhancing the cost-efficiency of electric vehicles.

- It is noticeable that the carbon footprint associated with the usage of an electric vehicle is not negligible. The origin of electricity utilized to operate an electric vehicle is derived from diverse sources, some of which exhibit a greater carbon footprint in comparison to others. Nonetheless, the carbon footprint associated with an electric vehicle remains considerably lesser in comparison to that of a vehicle powered by gasoline. Electric vehicles exhibit a reduced carbon footprint in comparison to their gasoline-powered counterparts. The reason for this is that electric vehicles are devoid of tailpipe emissions, whereas their gasoline-powered counterparts emit such pollutants.

- The energy efficiency and carbon footprint of sedans like the Tesla Model 3 and BMW i3 are better than those of bigger SUVs and trucks. That's because they're compact and light, two factors that contribute to this result. This is also evident from the greater carbon footprint of the SUV-style Audi e-tron compared to the sedan-style Tesla Model 3.

- SUVs have a significant carbon footprint per mile driven because of their greater energy consumption rate and longer recharge time compared to sedans and hatchbacks.

- Depending on the kind of vehicle model and driving behaviour, energy usage varies. For example, the sedan model was found to have a lower energy usage than the SUV type.

- Charging times for electric cars vary based on the area at which they were charged. Home charging facilities need more time than public charging stations.

All the visualisations generated above suggest that electric cars, especially those with greater energy efficiency and smaller carbon footprints, may be a viable option for lowering transportation-related carbon emissions in densely populated places like cities and suburbs. The total carbon footprint of electric cars may also be affected by factors like as the accessibility of charging infrastructure and the driving habits of individual users.

IV. CURRENT ADVANCEMENTS AND FUTURE PROJECTIONS

The use of artificial intelligence in environmentally friendly technologies is a fast expanding niche. By adjusting charging and discharging cycles in response to fluctuations in energy demand and supply, energy storage optimization helps to make the most of energy storage systems.

Leveraging AI and predictive analysis to EV manufacturing would have many positive effects. To begin, AI may be used to improve production by pointing out where improvements can be made and where bottlenecks exist. Second, manufacturers may modify output to meet demand without needing to keep extra units in store by using predictive research to project EV sales in the future. Finally, artificial intelligence and predictive analysis may be utilised to enhance EV quality by anticipating and preventing defects.

Advanced computation and predictions provide safer transportation for future generations. . AI and predictive analysis may improve electric car manufacturing efficiency and security. Modern computer systems' primary strength is their speed and precision in handling massive data sets. From raw material procurement through finished product quality control, producers may now examine and improve each and every step of the manufacturing process. To further ensure the manufacturing of high-quality electric cars and reduce the risk of errors, manufacturers may use predictive analysis to foresee future difficulties and change production procedures appropriately. Demand projections for EVs may also benefit greatly from the use of state-of-the-art computational technology. Overproduction or stockpiling of electric vehicles may be avoided with the use of forecasting technologies that analyse trends in consumer behaviour and external variables like government policies and laws.

Machine learning algorithms can examine real-time battery pack data to help build BMSs. AI can identify battery pack health and performance trends by evaluating temperature, voltage, and current data. This information may improve the battery pack's charging and discharging cycles, extending its lifespan and preventing unexpected failures.

Predictive maintenance strategies for EHVs can be created with the help of AI. Artificial intelligence (AI) can analyse battery performance and wear and tear data to identify early warning indications of possible faults and notify users and manufacturers so they may take preventative action. As a result, EHVs may be safer and more reliable, with less potential for unexpected failures or accidents.

With the proliferation of EHVs, transportation and environmental impact have both been greatly reduced. The potential of cyberattacks that might jeopardise the security and dependability of EHVs is rising, however, in tandem with their increasing popularity. Safe and trustworthy EHVs may be improved with the help of AI by bolstering their cyber defences.

Remote hijacking includes illegal access to the vehicle's electric control units(ECUs) through Bluetooth or Wi-Fi. A cyberattacker might manipulate the vehicle's acceleration and brakes, endangering the driver and passengers. EHVs may be remotely hijacked, and attackers can use communication protocol weaknesses to access ECUs. AI can monitor the vehicle's Bluetooth and Wi-Fi channels for signs of remote hijacking. Machine learning algorithms can discover security breaches in massive data sets. AI can detect and react to anomalies like abrupt acceleration or braking, which may indicate a remote hijacking attempt.

The Battery Management System (BMS) and infotainment system may be hacked to steal important data. Data breaches might disclose the car owner's personal information, jeopardise performance, and allow attackers to manipulate the vehicle's systems. EHVs' software and communication systems lack solid security standards, making them susceptible to data intrusions, according to researchers. The vehicle's onboard systems, may be monitored by AI to identify any unauthorised attempts to access or alter the vehicle's sensitive data. Unusual data access patterns, such as a spike in the number of read or write operations to the vehicle's storage devices, may be detected by machine learning algorithms that have been taught to do so. To prevent malicious malware from being installed in a car, AI may also be used to identify when software updates are attempted.

Illegal access entails physically accessing the BMS and ECUs and using specialised equipment to harvest sensitive data or manipulate the vehicle's software. This cybersecurity issue might undermine the vehicle's performance and steal critical data, such as the owner's personal information. EHVs lack physical protection, making them susceptible to illegal entry, according to research. AI may monitor the vehicle's BMS and ECUs for physical access. . AI can also identify efforts to remove sensors or introduce foreign things to the car.

Recent research has focused on sustainable electric transportation. Electric vehicle efficiency and sustainability must be improved as the world progresses towards a cleaner future. Integrating powerful AI and optimization methods into electric vehicle design and management is one of the best ways to achieve this aim. AI-based optimization may improve electric vehicle performance, sustainability, and environmental effect.Notably, quantum artificial intelligence has become a viable field of study for long-term electric vehicle sustainability. Complex optimization challenges in the design and administration of electric cars may be effectively addressed by using quantum computing's unprecedented processing capacity and computational speed. In order to reduce the environmental impact and maximise the longevity of electric cars, quantum optimization algorithms may be employed to optimise their energy usage while in operation.

In addition, both traditional AI and quantum AI may help with the creation of preventative maintenance strategies for electric cars. These strategies make use of machine learning algorithms to sift through mountains of data gathered by electric cars, including sensor readings, battery performance, and driving habits, in order to foresee and avoid breakdowns and other defects. By addressing maintenance issues before they become serious problems, like what was performed in particular for the methodological user case study, predictive maintenance systems have the potential to extend the life of electric vehicles, reduce repair costs, and enhance their environmental friendliness.

The convergence of AI, optimization, and quantum AI approaches provides a viable path to sustainable electric transportation. With these technologies, electric cars may be built and operated to maximise their performance, lessen their environmental effect, and improve their overall sustainability. Thus, study in this field may pave the path for a greener and more sustainable future. Using machine learning models to anticipate carbon footprints and energy consumption patterns may result in more precise estimates and give policymakers, producers, and consumers with important information.

Smart grids and charging facilities may help integrate electric cars into the power system, improving energy management. Electric cars may become more ecological and efficient as renewable energy sources like solar and wind become more available.

The development of driverless cars and the proliferation of ridesharing programmes provide new opportunities to increase the utility and effectiveness of electric vehicles. Energy consumption and greenhouse gas emissions may be lowered by optimising the charging and driving habits of electric cars using machine learning algorithms.

Integration of electric cars into a sustainable and efficient energy environment represents the future potential for this subject. The combination of technical advances and machine learning models may deliver insightful information and pave the path for a cleaner and greener future.

## V. CHALLENGES AND BOUNDATIONS

Despite its promise, green technology fueled by AI has a number of downsides. Consider the ethical and social repercussions of AI-powered energy systems, as well as the complexity of the energy data and knowledge gaps that exist. To overcome these issues, more investigation and development must be put into AI-powered green solutions, stakeholder participation, and proper AI utilisation in energy systems.

AI and electric vehicle research is hampered by data availability and dependability. It's hard to get reliable and complete statistics on electric car use, charging habits, and emissions across driving patterns and locales. The research may also be limited by the quantity and kind of electric cars available for investigation or the geographic region in which it may be done.

AI models used to evaluate data must be accurate and fair. For accurate predictions, models must be trained and evaluated. The models must also be transparent and interpretable so researchers and stakeholders can understand how and why they make predictions. There may be significant computational requirements for running AI models. Such resources may include powerful computers, large amounts of storage space, and plenty of RAM.

The problem of inconsistent data standards or lack of standardization across sources is another difficulty. It may be challenging to compare and evaluate the findings if the data were obtained using various methodologies and units. Interpreting and communicating study findings may be difficult. It's crucial to convey findings simply and properly both specialists and non-experts.

Lastly, ethical and regulatory concerns might require additional considerations while studying the intersection of AI and EVs. Collecting and analysing data on people's driving and charging habits, for instance, might raise privacy issues. It's also possible that gaining participant permission and protecting the privacy of any data obtained are legally mandated steps that must be taken.

In spite of these barriers, AI and electric vehicle research offers enormous potential for reducing greenhouse gas emissions and mitigating the effects of climate change. By creating precise models and tools for forecasting and optimising electric vehicle usage and charging patterns, experts may contribute to the advancement of the transition to a more sustainable and low-carbon transportation system.

## VI. CONCLUSION

Electric cars had a less carbon impact in suburban and rural travel patterns, according to the research. The report also demonstrated that machine learning algorithms have the ability to provide reliable forecasts of EVs' energy use and carbon emissions.

These studies imply electric cars might reduce transportation's environmental effect. Electric cars emit less carbon than gasoline vehicles, and energy-efficient driving may cut emissions even more. Nevertheless, improvements in battery technology and charging infrastructure are still required in order to completely fulfil the promise of electric cars.

The findings of this study provide a basic framework for more research into improving the efficiency of and reducing the environmental impact of electric vehicles. Obtaining high-quality data in a standardised format is yet another concern. These issues may be explored further in future studies, as can the possibility of widening the scope of study to include more vehicle types and environmental elements. Despite this, electric cars have the potential to play a significant role in the transportation industry of the future if sufficient research and development efforts are made.

## ACKNOWLEDGMENT

I would like to express my gratitude to individuals who have aided me with their respective feedbacks for the research. I'd also like to thank the people who took part in this study and gave us their valuable information. Their assistance is deeply appreciated. I am thankful to the editorial board and peer reviewers for their insightful comments and suggestions that assisted me in refining the research findings. The paper's effectiveness has been substantially enhanced by their insightful critique and recommendations.